\documentclass[10pt, a4paper]{article}

\usepackage{lrec-coling2024} %
\usepackage{xcolor}
\usepackage{graphicx}
\usepackage{tabularx}
\usepackage{enumitem}
\usepackage{booktabs}
\usepackage{multirow}
\usepackage{float}

\definecolor{neonyellow}{RGB}{204,255,0} %

\title{M\textsuperscript{3}TCM: Multi-modal Multi-task Context Model for Utterance Classification in Motivational Interviews}

\name{Sayed Muddashir Hossain, Jan Alexandersson, Philipp M\"uller}

\address{ 
    German Research Center for Artificial Intelligence\\ %
    Saarbr\"ucken, Germany \\ %
     sayed\_muddashir.hossain@dfki.de, jan.alexandersson@dfki.de, philipp.mueller@dfki.de\\
}

\abstract{
Accurate utterance classification in motivational interviews is crucial to automatically understand the quality and dynamics of client-therapist interaction, and it can serve as a key input for systems mediating such interactions. Motivational interviews exhibit three important characteristics. First, there are two distinct roles, namely client and therapist. Second, they are often highly emotionally charged, which can be expressed both in text and in prosody.
Finally, context is of central importance to classify any given utterance. Previous works did not adequately incorporate all of these characteristics into utterance classification approaches for mental health dialogues. In contrast, we present M\textsuperscript{3}TCM, a \underline{M}ulti-\underline{m}odal, \underline{M}ulti-\underline{t}ask \underline{C}ontext \underline{M}odel for utterance classification. Our approach for the first time employs multi-task learning to effectively model both joint and individual components of therapist and client behaviour. Furthermore, M\textsuperscript{3}TCM integrates information from the text and speech modality as well as the conversation context.
With our novel approach, we outperform the state of the art for utterance classification on the recently introduced AnnoMI dataset with a relative improvement of 20\% for the client- and by 15\% for therapist utterance classification. In extensive ablation studies, we quantify the improvement resulting from each contribution.
 \\ \newline  }

\begin{document}

\maketitleabstract

\section{Introduction}

Motivational interviewing (MI) is an important tool in helping clients to achieve goals such as reducing alcohol consumption and smoking, managing asthma or diabetes, or increasing physical activity.
Automatic analysis of motivational interviewing has on the one hand the potential to improve our understanding of the effectiveness of different techniques.
On the other hand, it is also a basis for building social agents that can meaningfully interact with clients.
To this end, automatic approaches need to be able to precisely categorize the utterances of both counselor and client.

Motivational interviews have three important characteristics.
First, client and therapist have distinct roles, including different sets of ground truth utterance labels~\cite{wu2022anno}.
Second, motivational interviews are often emotionally charged.
Third, conversation context is crucial to interpret any given utterance.
Previous approaches to utterance classification in motivational interviews did not fully take advantage of all of these characteristics. 
While approaches integrating text and audio do exist~\cite{aswamenakul2018,gupta2014,singla2018,tavabi2020}, they commonly do not model the conversation context.
The few approaches that do model conversation context, are either not multi-modal~\cite{tavabi2020}, or only considered a single utterance as context~\cite{gupta2014}.
Most importantly, all previous approaches addressed patient and therapist utterance classification in completely separate models.
The potential benefit of multi-task learning remains unexplored.

To overcome these limitations, we present M\textsuperscript{3}TCM, a multi-modal multi-task context model for utterance classification in motivational interviewing. 
M\textsuperscript{3}TCM for the first time uses multi-task learning to effectively model both joint and individual components of the two tasks of classifying therapists' and clients' utterances. 
Our approach furthermore effectively leverages prosodic information as well as the conversation context.
In evaluations on the recently introduced AnnoMI dataset~\cite{wu2022anno}, M\textsuperscript{3}TCM outperforms previously proposed approaches by a significant margin (0.66 F1 vs. 0.55 F1 for client utterances, 0.83 vs. 0.72 F1 for therapist utterances).
We present extensive ablation experiments, documenting the importance of the multi-task framework and of utilizing text and audio modalities in conjunction with conversation context.
We furthermore for the first time evaluate different sizes of the input window, showing that the optimal context size is significantly larger than those used in previous work.
\section{Related Work}
Our work is related to utterance classification in mental health conversations and to multi-task learning approaches applied to conversation analysis.

\subsection{Utterance Classification in Mental Health Conversations}
\citet{ewbank2020quantifying} classified therapist utterances obtained from transcripts of Cognitive Behaviour Therapy~\cite{brewin2006understanding} sessions into 24 categories to predict therapy outcome.
In another study, \citet{ewbank2021understanding} employed deep learning techniques to automatically classify patient talk types within Cognitive Behaviour Therapy. 

Previous approaches confirm the importance of fusing text and audio information for utterance classification in MI~\cite{aswamenakul2018,tavabi2020,gupta2014,singla2018}.
Most approaches only address the problem of client talk type classification, but \citet{singla2018} proposed an approach based on single utterances that is applied to therapist and client talk type classification, integrating text and audio information. 
Only a subset of utterance classification approaches modeled the conversation context in order to classify a target utterance.
\citet{tavabi2020} took 3 previous text-utterances from both client and therapist as context to classify current client utterance.  
\citet{gupta2014} investigated the effect of laughter and prosodic differences in MI interviews, using the previous therapist's utterance as context. 
In summary, while several approaches integrated text and audio modalities, these commonly do not explore the effect of the size of the context.
The few approaches that do model conversation context do not provide analyses on the impact of the size of the input window.
Crucially, none of the existing approaches leverages a multi-task learning framework to simultaneously learn models for therapist and client utterances.

Recently \citet{wu2022anno} introduced AnnoMI, an expert-annotated dataset of motivational interviews available on Youtube.
The dataset derives its annotations from the Motivational Interviewing Skills Code (MISC)~\cite{miller2012motivational} and has a different set of labels for client and therapist. We use AnnoMI because it is the biggest publicly available dataset with MI interviews, annotated by experts.
Existing work on this dataset employed language models to create separate, single-utterance text-based classifiers for therapist and client utterances~\cite{wu2022towards, fi15030110}.
To the best of our knowledge, we present the first multi-modal, context-aware, multi-task approach to utterance classification on the AnnoMI corpus.

\subsection{Multi-Task Approaches}
Previous work applied multi-task learning for dialogue analysis in several setups.
\citet{ide2021multi} proposed a multi-task learning method for emotion-aware dialogue response generation, emphasizing the synergy between generation and classification tasks. They train the same model to generate dialogue responses and at the same time detect emotion.
\citet{liu2022emodm} introduced EmoDM, which at the same time learns to track emotional states and empathetic dialogue policy selection. 
\citet{kollias2022abaw2} presented the ABAW Competition, which includes challenges like Valence-Arousal Estimation and Expression Classification using multi-task learning on the Aff-Wild2 database.
In a subsequent iteration, \citet{kollias2022abaw2} highlighted the potential of multi-task approaches in emotion detection and classification using both synthetic data and multi-task learning to classify valance or arousal and emotions.
To the best of our knowledge, multi-task learning was not yet applied to model the different roles speakers have in motivational interviewing.

\section{Method}
\begin{figure*}[h]
    \centering
    \includegraphics[width=\linewidth]{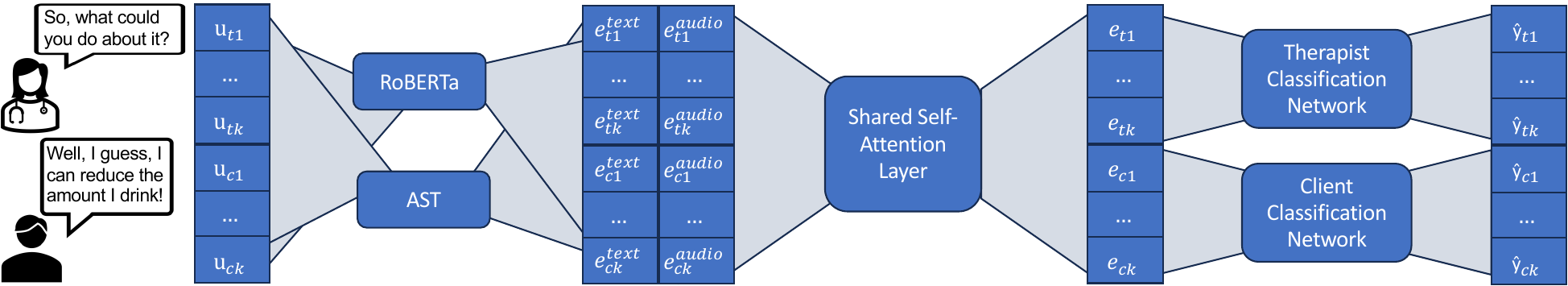}
    \caption{Overview over the M\textsuperscript{3}TCM Model. Several consecutive therapist and client utterances ($u_{ti}$ and $u_{ci}$, respectively) are encoded using RoBERTa and AST models, producing text and audio embeddings. A shared self-attention layer models conversation context across utterances. Finally, separate classification networks produce predictions for therapist and client utterances.} 
    \label{fig:multimodalmtl}
\end{figure*}

\autoref{fig:multimodalmtl} illustrates the architecture of our model. 
Text- and audio embeddings are extracted from $k$ consecutive utterances of therapist and client.
A shared self-attention layer is used to model conversation context across utterances, and task-specific classification networks are utilized to produce classification outputs for the therapist and client.

\subsection{Input Embeddings}

In the following we discuss how we obtained per-utterance embeddings from text and audio inputs.

For text data, we used RoBERTa Large~\cite{liu2019roberta}. RoBERTa, short for "Robustly optimized BERT approach," is a variant of the BERT model designed for natural language processing. 
RoBERTa improved BERT's performance by altering the training regimen, notably removing the next-sentence prediction objective and utilizing dynamic masking for more efficient pre-training. 
The model was trained with more data and larger batch sizes, resulting in improved accuracy and demonstrating the significance of meticulous training details. 
RoBERTa achieved state-of-the-art results in various NLP benchmarks~\cite{liu2019roberta} including emotion~\cite{adoma2020comparative} and depression~\cite{gupta2023comparative} recognition.

To encode prosodic information, we made use of the Audio Spectrogram Transformer (AST)~\cite{gong2021ast}, a specialized model designed to handle audio classification tasks using the transformer architecture. AST directly operates on audio spectrograms and achieved state-of-the-art results
on recognizing human speech~\cite{gemmeke2017audioset}, command~\cite{warden2018speech} and also difference between human and environmental sounds~\cite{piczak2015esc}. 
The Audio Spectrogram Transformer (AST)~\cite{gong2021ast} is particularly well-suited for analysing prosody due to its ability to model inter-dependencies across time and thereby extract intricate patterns from audio data. AST's attention-based mechanism allows it to focus on specific aspects of the audio spectrogram, such as the variations in pitch, tempo, and volume, which are integral components of prosody.

\subsection{M\textsuperscript{3}TCM Approach}

Our model processes $k$ utterances of each patient and therapist in parallel.
At each time step $i=1..k$ we have the textual utterances of the therapist and client, denoted as $u_{ti}$ and $u_{ci}$ respectively, along with their corresponding audio spectrograms.
RoBERTa is used to extract $2k$ text embeddings $E_{text} = e^{text}_{t1}..e^{text}_{tk},e^{text}_{c1}..e^{text}_{ck}$ from therapist and client utterances.
AST on the other hand produces the corresponding audio embeddings $E_{audio} = e^{audio}_{t1}..e^{audio}_{tk},e^{audio}_{c1}..e^{audio}_{ck}$.

Text and audio embeddings for any given utterance are subsequently concatenated:
\begin{equation}
E = E_{audio} \oplus E_{text}
\end{equation}
which results in the combined embeddings $E = e^{}_{t1}..e^{}_{tk},e^{}_{c1}..e^{}_{ck}$. 
To incorporate conversation context in our classification approach, we model relations between utterances with a self-attention layer~\cite{vaswani2017attention}:
\begin{equation}
E' = SelfAttention(E)
\end{equation}
Our multi-task learning approach employs two task-specific networks working on separate subsets of the $2k$ transformed embeddings $E'=e'_{t1}..e'_{tk},e'_{c1}..e'_{ck}$.
The therapist utterance classification network $f_{t}$ receives as input the first $k$ embeddings $E'_{therapist}=e'_{t1}..e'_{tk}$ and outputs $k$ classification decisions $\hat{y_t}=\hat{y}_{t1}..\hat{y}_{tk}$, one for each therapist input utterance:
\begin{equation}
\hat{y_t} = f_{t}(E'_{therapist})
\end{equation}
Analogously, the client classification network $f_c$ produces predictions $\hat{y}_c = \hat{y}_{c1}..\hat{y}_{ck}$ for the $k$ client input utterances.
While client and therapist classification networks have separate weights, the self-attention layer is shared between both tasks.
This allows our self-attention layer to learn both task-dependent and task-independent aspects of behaviour.
To be precise, the multi-task-learning takes place through query-key interactions across client and therapist utterances.

\subsection{Implementation Details}
To address the class imbalances(Sub-section \ref{sec:Dataset}) resulting from client and therapist behaviour, we use the Focal Loss function, suitable for imbalanced classification scenarios \citet{lin2017focal}.  
M\textsuperscript{3}TCM Shared layer has the dimension of \( 1551 \times 1024\). Both client and therapist specific heads have two layers, with \( 1024 \times 512\) and \( 512 \times 256\) dimensions.
To improve reproducibility, we make our code publicly available\footnote{\url{https://git.opendfki.de/philipp.mueller/m3tcm}}.

\begin{table*}[t]
\centering
\vspace{10pt}
\resizebox{\textwidth}{!}{%
\begin{tabular}{lcccccccccc}
\toprule
& \multicolumn{5}{c}{Client} & \multicolumn{5}{c}{Therapist} \\
\cmidrule(lr){2-6} \cmidrule(lr){7-11}
Models & Average & Change & Neutral & Sustain & & Average & Reflection & Question & Input & Other \\
\midrule
Random Baseline &  0.33 & 0.25 & 0.63 & 0.12 & & 0.25 & 0.25 & 0.29 & 0.15 & 0.31 \\
\citet{fi15030110}& 0.55 & 0.51 & \textbf{0.74} & 0.39 & & 0.72 & 0.77 & 0.86 & 0.63 & 0.64 \\
\midrule
M\textsuperscript{3}TCM Without Finetuning & 0.54 &  0.70 & 0.42 & 0.41 & & 0.73 & 0.65 & 0.82 & 0.81 & 0.63 \\
M\textsuperscript{3}TCM Text Only Single Task & 0.58 & 0.76 & 0.56 & 0.43 & & 0.77 & 0.73 & 0.86 & 0.82 & 0.68 \\
M\textsuperscript{3}TCM Audio Only Single Task & 0.40 & 0.65 & 0.38 & 0.18 & & 0.44 & 0.40 & 0.60 & 0.44 & 0.31 \\
M\textsuperscript{3}TCM Audio Only No Context & 0.38 & 0.65 & 0.36 & 0.13 & & 0.40 & 0.38 & 0.58 & 0.40 & 0.25 \\
M\textsuperscript{3}TCM Text Only No Context & 0.57 &  0.73 & 0.52 & 0.45 & & 0.77 & 0.74 & 0.86 & 0.82 & 0.67 \\
M\textsuperscript{3}TCM Audio Only & 0.46 & 0.73 & 0.43 & 0.21 & & 0.49 & 0.46 & 0.68 & 0.48 & 0.33 \\
M\textsuperscript{3}TCM Text Only & 0.63 &  0.80 & 0.59 & 0.49 & & 0.80 & 0.76 & 0.89 & 0.85 & 0.68 \\
M\textsuperscript{3}TCM No Context & 0.61 &  0.78 & 0.57 & 0.48 & & 0.76 & 0.70 & 0.83 & 0.87 & 0.65 \\
M\textsuperscript{3}TCM Single Task & 0.60 &  0.78 & 0.57 & 0.46 & & 0.77 & 0.70 & 0.85 & 0.87 & 0.65 \\
\midrule
M\textsuperscript{3}TCM & \textbf{0.66} &  \textbf{0.83} & 0.62 & \textbf{0.52} & & \textbf{0.83} & \textbf{0.81} & \textbf{0.89} & \textbf{0.88} & \textbf{0.73} \\
\bottomrule
\end{tabular}
}
\caption{Classification results for M\textsuperscript{3}TCM compared to baselines and ablation conditions. We report per-class, as well as macro-averaged F1 scores for both client and therapist classification tasks.}
\label{tab:our_results}
\end{table*}

\section{Experiments}

\subsection{Data Prepossessing}
\label{sec:Dataset}
We began with the AnnoMI dataset from \citet{fi15030110}, consisting of 13551 utterances transcribed from 133 Youtube videos. 
Since the initial publishing of AnnoMI, some of those videos have been removed from Youtube, our dataset contained in the end 125 videos.
Given our multi-modal approach including audio, we had to remove utterances of non-available videos, leaving us with 12778 instances, 6338 for client and 6440 for therapist. 
To extract the per-utterance audios, we isolated audio from videos and segmented them using the utterance timestamps provided with AnnoMI. 
Instances with multiple annotators were harmonized by selecting the most frequent annotation.

Our targets were the client talk type class and the main therapist behaviour. 
The client class was imbalanced: 63\% ``neutral'', 25\% ``change'', and 12\% ``sustain''. On the other hand, the therapist's class distribution showcased a more even spread: 31\% for ``other'', 29\% for ``question'', 25\% for ``reflection'', and 15\% for ``therapist\_input''. \

\subsection{Training Details}
We used 5 Fold Cross Validation stratified by video to guarantee that no utterances form the same video can appear both in train and test sets.
We used $\frac{3}{5}$ of the data for training, $\frac{1}{5}$ for validation, and $\frac{1}{5}$ for testing.

In a first step, we fine tuned both AST~\cite{gong2021ast} and RoBERTA Large~\cite{liu2019roberta} model on our dataset. 
We also tried using AST and RoBERTa without finetuning, but that led to inferior results.
In a second step, we trained the full M\textsuperscript{3}TCM model for 100 epochs and choose the best model based on the performance on the validation set. One thing to note is that at this stage of the training the weights of the finetuned RoBERTa and AST layer were fixed and as we said before was selected based on the best performance on the validation set.

Both for the finetuning and the final training phase we used the the AdamW optimizer at a learning rate of 1e-5 \citet{loshchilov2017decoupled} and trained for 100 epochs.
We selected the best model from these 100 epochs by evaluating F1 score on the validation set.

\section{Results}
In line with previous work~\cite{wu2022towards, fi15030110}, we evaluated all approaches using the F1 score.
We do so both with per-class F1 scores as well as separate macro-averaged F1 scores for the patient and therapist utterance classification tasks.
In ~\autoref{tab:our_results}, we report results for M\textsuperscript{3}TCM as well as baselines and ablation conditions.

M\textsuperscript{3}TCM outperforms all other approaches, reaching 0.66 F1 for the client and 0.83 F1 for therapist utterance classification.
This is a clear improvement over the previous state of the art by \citet{fi15030110} (0.55 F1 client, 0.72 F1 therapist).
Crucially, our ablation experiments confirm the utility of multi-task learning.
Models trained separately on patient and therapist utterance classification (``M\textsuperscript{3}TCM Single Task''), only reached 0.60 F1 for client and 0.77 F1 for therapist.
Furthermore, consistent improvements over mono-modal ablations (``M\textsuperscript{3}TCM Audio Only / Text Only'') document the utility of fusing text and audio information.
In addition, we observed that the inclusion of conversation context leads to clear improvements: we see that without context it achieves for client F1 of 0.61 and for therapist 0.76.

\begin{figure}[t]
    \centering
    \includegraphics[width=\columnwidth]{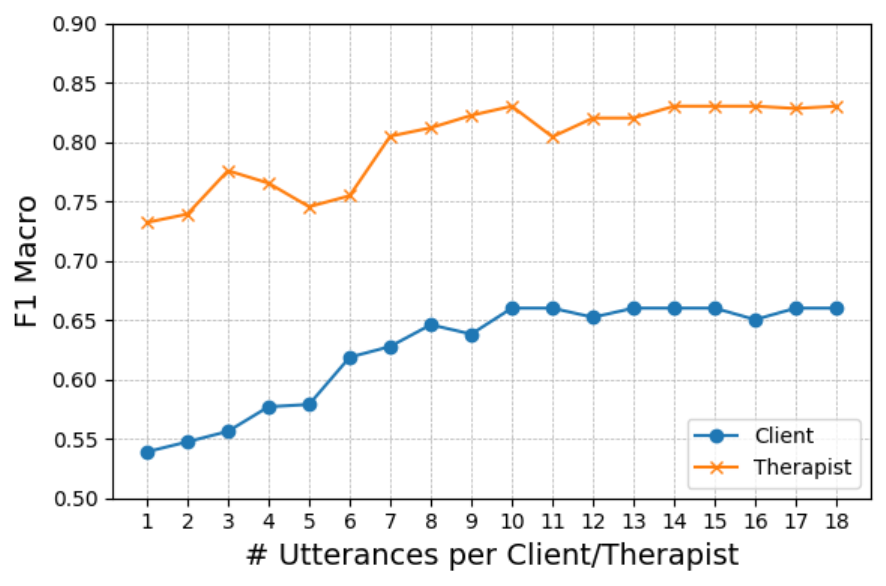}
    \caption{Performance for therapist and client utterance classification for different context sizes.}
    \label{fig:dialogue_window}
\end{figure}

Our M\textsuperscript{3}TCM model has a slightly lower F1 score (0.62) for the majority "neutral" class for client talk type compared to random guessing (0.63 F1). The reason for this is that we decided to optimise our model to perform well on all classes (and not primarily on the majority class), which is reflected in consistently higher scores for the minority classes. For “change”, M\textsuperscript{3}TCM reached 0.83 F1 versus 0.51 F1 for \citet{wu2022towards}, and 0.25 F1 for the random baseline.
For the challenging minority class “sustain”, M\textsuperscript{3}TCM reached 0.52 F1 versus 0.39 F1 for \citet{wu2022towards}, and 0.12 F1 for the random baseline. 
A precise distinction between “change” and “sustain” is especially important in motivational interviews, as these are highly informative classes concerning behaviour change.

To better understand the utility of conversation context, we conducted an experiment with varying numbers of utterances as context (see \autoref{fig:dialogue_window}). 
We observed a clear increase in F1 score for both therapist and patient when increasing the number of patient/therapist utterances we input to the model from 1 to 10.
For more than 10 utterances performance reaches a plateau, while memory utilization continues to increase.
We therefore determine 10 utterances per patient/therapist as the optimal input size, which is much larger than the 
input window of maximally 3 utterances used in previous work~\cite{tavabi2020}.

\begin{figure}[t]
    \centering
    \includegraphics[width=\columnwidth]{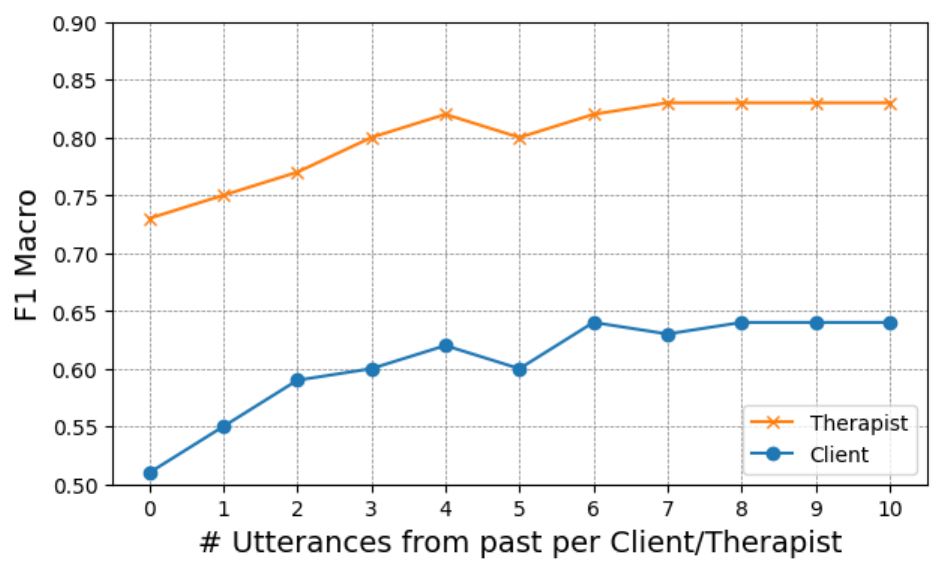}
    \caption{Performance of therapist and client utterance classification for different context sizes in an online evaluation scenario.} 
    \label{fig:dialogue_window_online_model}
\end{figure}

It is important to note that our model is evaluated in an offline scenario, i.e. for the classification of a given utterance it also has access to future utterances.
To understand its capabilities in an online classification setup, we analyze the prediction performance when only using the prediction on the last utterance of the input window.
We present the corresponding results for varying sizes of the input window (i.e. previous) utterances in \autoref{fig:dialogue_window_online_model}.
In general, the performance is very similar to the offline approach, demonstrating the utility of our approach in online classification scenarios.

\section{Conclusion and Future Work}
In this work, we presented M\textsuperscript{3}TCM, a multi-modal and context-sensitive approach to utterance classification in motivational interviews that for the first time leverages multi-task learning to model both therapist and patient at the same time.
We showed clear improvements over the previous state of the art as well as ablated versions of our model.
As such, our work underlines the importance of models that make use of all the available information to build highly accurate conversation analysis systems.
For future work, it would be interesting to integrate the video modality alongside text and prosody.
Furthermore, our multi-task approach could be applied to different scenarios that exhibit asymmetrical roles in conversation.
These may include psychiatric interactions~\cite{konig2022multimodal}, sales conversations, teacher-student interactions~\cite{cafaro2017noxi}, or police interrogations.
In addition, it will be interesting to integrate predicted utterance classes as input features in nonverbal conversational behaviour generation approaches~\cite{withanage2023renelib}.

\section{Acknowledgements}
J. Alexandersson and P. M\"uller were funded by the European Union Horizon Europe programme, grant number 101078950.

\nocite{*}
\section{Bibliographical References}\label{sec:reference}

\bibliographystyle{lrec-coling2024-natbib}
\bibliography{lrec-coling2024-example}

\end{document}